\documentclass[10pt,twocolumn,letterpaper]{article}

\usepackage{iccv}
\makeatletter
\@namedef{ver@everyshi.sty}{}
\makeatother
\usepackage{times}
\usepackage{epsfig}
\usepackage{graphicx}
\usepackage{amsmath}
\usepackage{amssymb}
\usepackage{subfig}
\usepackage{floatrow}
\usepackage{booktabs}
\usepackage{enumitem}

\iccvfinalcopy 

\def\iccvPaperID{3750} 

\ificcvfinal\fi
\begin{document}

\title{Face De-occlusion using 3D Morphable Model and Generative Adversarial Network}

\author{Xiaowei Yuan and In Kyu Park\\
{\tt\small \{xiaoweichn@qq.com~~ pik@inha.ac.kr\} }\\
Dept. of Information and Communication Engineering, Inha University, Incheon 22212, Korea\\
}

\maketitle

\begin{abstract}
In recent decades, 3D morphable model (3DMM) has been commonly used in image-based photorealistic 3D face reconstruction. However, face images are often corrupted by serious occlusion by non-face objects including eyeglasses, masks, and hands. Such objects block the correct capture of landmarks and shading information. Therefore, the reconstructed 3D face model is hardly reusable. In this paper, a novel method is proposed to restore de-occluded face images based on inverse use of 3DMM and generative adversarial network. We utilize the 3DMM prior to the proposed adversarial network and combine a global and local adversarial convolutional neural network to learn face de-occlusion model. The 3DMM serves not only as geometric prior but also proposes the face region for the local discriminator. Experiment results confirm the effectiveness and robustness of the proposed algorithm in removing challenging types of occlusions with various head poses and illumination. Furthermore, the proposed method reconstructs the correct 3D face model with de-occluded textures.
\end{abstract}

\vspace*{-2mm}
\section{Introduction}
3D face reconstruction from a single image is a key technology in many computer vision and graphic applications, such as face recognition and face animation. Since Blanz and Vetter~\cite{Blanz_1999} proposed the 3D morphable face model~(3DMM), the methodology based on 3DMM has been most popular for coarse face geometric reconstruction. Further development involves using the shape from shading~(SfS) technique to enhance details ({\em e.g.} wrinkles) on the face geometry~\cite{Kemelmacher_2011, Richardson_2017_CVPR, Jiang_2018}. These techniques assume that the input image is free from occlusion, or at most, self-occluded by head pose variation. However, in actual situations in the wild, we encounter new challenges in which existing algorithms become inapplicable due to serious occlusion by eyeglasses, masks, hands, and others.

\begin{figure}
\begin{center}
{\includegraphics[width=0.24\columnwidth]{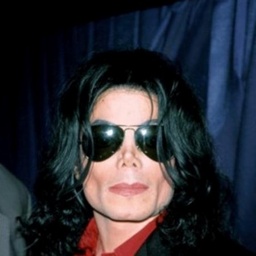}}~%
{\includegraphics[width=0.24\columnwidth]{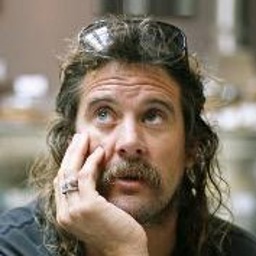}}~%
{\includegraphics[width=0.24\columnwidth]{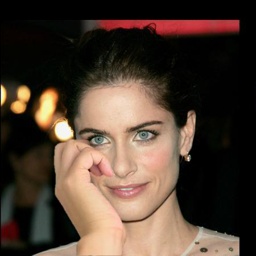}}
{\includegraphics[width=0.24\columnwidth]{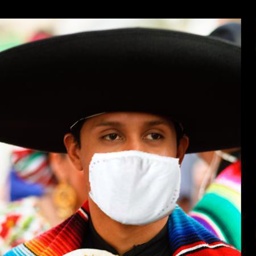}}~%
\\ \vspace{-3mm}
\subfloat[]{\includegraphics[width=0.24\columnwidth]{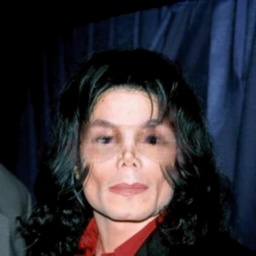}}~%
\subfloat[]{\includegraphics[width=0.24\columnwidth]{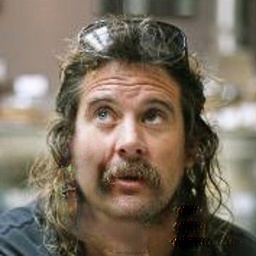}}~%
\subfloat[]{\includegraphics[width=0.24\columnwidth]{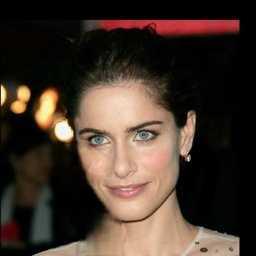}}
\subfloat[]{\includegraphics[width=0.24\columnwidth]{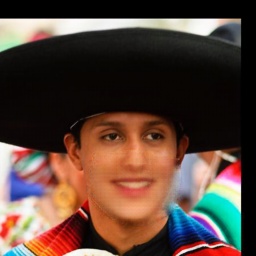}}~%
\end{center}
\vspace{-5mm}
\caption{Face de-occlusion results by applying the proposed method. (a)(b) Real images. (c)(d) Synthetic images. (Upper) Input images with occlusions. (Lower) De-occlusion results.}
 \label{fig:ResultIntro}
\end{figure}

To solve this problem in face recognition, a few methods propose solutions for automatic face de-occlusion to improve recognition performance~\cite{Cheng_2015, Jeong_2005, Wring_2009, Zhao_2018, Liu18}. However, almost all existing methods work under highly constrained conditions, {\em e.g.} low-resolution grayscale images with predefined head pose. Therefore, these methods cannot perform photorealistic 3D face reconstruction in terms of image resolution and image diversity.

The recent work by Tran~\cite{Tran_2018_CVPR} addresses the challenge of detailed face reconstruction from occluded images. However, rather than performing de-occlusion, this method focuses on geometrical reconstruction by searching the reference dataset to reconstruct the bump map on the occluded region. By contrast, our proposed method directly removes the occlusion on the face image, thereby allowing texture mapping on the reconstructed 3D model.

In this paper, we address the problem of face de-occlusion from seriously occluded actual images while focusing on the application of 3D face reconstruction. The proposed method aims to directly remove occlusions, thereby enabling it to synthesize the texture for the 3D face model. An example of face de-occlusion is shown in Figure~\ref{fig:ResultIntro}. Various occlusions and head poses are observed from actual face images. Thus, automatically removing face occlusions through a purely data-driven manner is a challenging task.

To solve this problem, we propose a novel 3DMM-conditioned deep convolutional neural network to learn to automatically remove occlusions. To the best of our knowledge, this is the first face de-occlusion network that attempts to exploit the potential use of 3DMM for face de-occlusion. Similar with the previous works~\cite{Deng_2018_CVPR, Li_2017_CVPR, Yin_2017_ICCV}, we employ the recent generative adversarial network (GAN)~\cite{NIPS2014_5423} that has been widely used to train an image synthesis model with strong ability to generate natural and high-quality images. In the proposed approach, global and local discriminators are combined with the generative model to achieve high-quality image synthesis. During training, 3DMM not only serves as prior but also proposes face region for the local discriminator. To diversify the training images with various occlusions, we synthesize a large-scale dataset from 300W-3D and AFLW2000-3D~\cite{Zhu_2016_CVPR}. The key contributions of this paper can be summarized as follows:
\begin{itemize}
  \item We propose a novel deep face de-occlusion framework, which applies the inverse use of 3DMM and GAN and consists of a generator and two discriminators.
  \item The proposed face de-occlusion model can handle face images under challenging conditions, {\em e.g.}, serious occlusions with nontrivial head poses and illumination variations.
  \item We build a large-scale synthesized face-with-occlusion dataset. All occlusions are semantically placed on the face with reference to face landmarks.
  \item The proposed face de-occlusion method not only boosts the performance of 3D face reconstruction but also allows face attribute editing by modifying the 3DMM coefficients.
\end{itemize}



\section{Related Works}


\paragraph{Image Completion}
Image completion or image inpainting aims to recover masked or missing regions on images with visually plausible contents. Recently, the generative model has been widely used in image completion with reasonably acceptable results~\cite{Iizuka_2017, Deng_2018_CVPR, Li_2017_CVPR, Song_2018, Yu_2018_CVPR}. These methods train an auto-encoder to predict the missing region by using a combination of reconstruction loss and adversarial loss. Despite the ability of the image completion technique to recover high-quality visual patterns in face de-occlusion tasks, the occluded region needs to be masked manually or with an additional object detection algorithm to segment the occluder. By contrast, the proposed model does not need any preprocessing on the occluded region and can automatically remove the occlusion.

\vspace*{-3mm}
\paragraph{Face De-occlusion and Frontalization}
Conventional face de-occlusion algorithms are developed to increase the performance of face recognition algorithms. Wright~{\em et al.}~\cite{Wring_2009} proposed to apply sparse representation to encode faces and demonstrated the robustness of the extracted features to occlusion. Cheng~{\em et al.}~\cite{Cheng_2015} introduced the double-channel SSDA~(DC-SSDA) to detect noise by exploiting the difference between activations of two channels. Recently, a deep learning-based approach has been proposed by Zhao~\cite{Zhao_2018} to restore the partially occluded face in several successive processes using an LSTM auto-encoder.

However, these methods can only remove occlusions under constrained conditions. Images in low-resolution grayscale and all faces in the dataset have to be cropped and aligned first. Therefore, these methods cannot conduct practical applications beyond face recognition. By contrast, the proposed method is targeted to perform actual face reconstruction and texture synthesis. Thus, we generalize the input to RGB images with enlarged resolution (256$\times$256) and with various head poses.

Besides, Yin \etal proposed a deep 3DMM-conditioned face frontalization method called FF–GAN~\cite{Yin_2017_ICCV}, which incorporates 3DMM coefficients into the GAN structure to provide poses prior to the face frontalization task. This method utilizes 3DMM coefficients as a weak prior to reduce the artifacts during frontalization in extreme profile views.

\vspace*{-4mm}
\paragraph{3D Face Reconstruction from Occluded Image}
Bernhard~{\em et al.}~\cite{Egger2018} proposed an occlusion-aware face modeling method in which they incorporated 3DMM as appearance prior in a RANSAC-like algorithm. In this method, the input image is segmented into face and non-face regions and the illumination is estimated using the face region only. However, this method is not robust if the occlusions, e.g., hands, have a similar color appearance to the face. A recent face alignment technique shows the robustness in occluded face images~\cite{Bulat_2017_ICCV, Xiong_2013_CVPR}. Thus, this technique can be used to fit 3DMM and produce a 3D face model with a few details. Although their goal is to robustly find the head pose, de-occlusion in not within the scope of their work. Tran~{\em et al.}~\cite{Tran_2018_CVPR} is the first to address the problem of detailed face reconstruction from occluded images by filling in the corrupted region of the bump map using a similar patch in a reference dataset. Although this method can generate a complete representation of face details, the de-occluded face image is not reconstructed~\cite{Tran_2018_CVPR}.

\begin{figure*}[t]
\begin{center}
    {\includegraphics[width=1.0\linewidth,keepaspectratio]{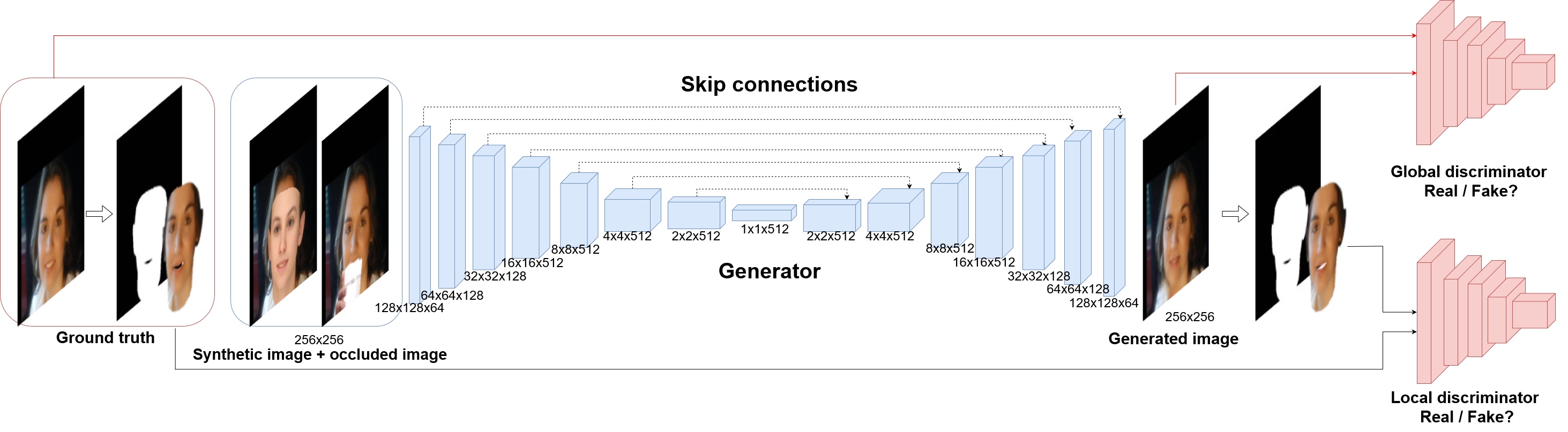}}
\end{center}
\vspace{-3.6mm}
\caption{Proposed network structure. It consists of a generator with two discriminators. The generator takes a synthesis image and an occluded image as input. Two discriminators help to generate a more natural result. Only the generator is necessary during testing.}
    \label{fig:Network_Structure}
\end{figure*}
\vspace*{-4mm}
\paragraph{Face Synthesis with GAN}
GAN~\cite{NIPS2014_5423} utilizes min–max optimization over the generator and discriminator and shows significant improvement in face synthesis applications, such as face attribute editing~\cite{Shen_2017}, and face completion~\cite{Deng_2018_CVPR, Li_2017_CVPR}. Gecer~{\em et al.} exploited to synthesize facial images~\cite{Gecer_2018} and facial textures~\cite{Gecer_2019} conditioned on latent 3DMM parameters. However, no previous study has been conducted on using GAN for de-occlusion on challenging faces.
\section{Overview and Background}
\subsection{Overview of the Proposed Method}
In this paper, we address the face de-occlusion problem by incorporating 3DMM and GAN in the same framework. Motivated by~\cite{Yin_2017_ICCV}, we propose to use 3DMM as our geometric regularization in our face de-occlusion model. In our work, 3DMM is a strong prior without which the algorithm would fail completely. That is, 3DMM is used to provide constraints on the appearance of the occluded region in which the generator output is used explicitly to synthesize the de-occluded image.

We first fit the 3DMM to an occluded image and synthesize a 2D face image. Then, we take the synthesis and occluded face as the inputs of the generator to synthesize occlusion-free images. At the same time, a global discriminator and a local discriminator attempt to distinguish the image as a real image or a generated one. The 3DMM serves not only as the geometric prior but also provides the face region for the local discriminator. With the guidance of 3DMM, the generator can efficiently remove occlusions even on challenging faces. Figure~\ref{fig:Network_Structure} illustrates our proposed framework consisting of a 3DMM-conditioned generator and two discriminators.

\subsection{3D Morphable Model}
3DMM is the most commonly used statistical method for the representation and synthesis of face geometry and texture. In our work, we use a multilinear 3DMM with 53K vertices and 106K triangles to represent the 3D face shape~\cite{xiangyu_2015}.
Each face geometry can be parameterized as follows:
\begin{equation}
  {M}(\boldsymbol{\alpha,\beta})=\boldsymbol{\bar S_{id}}+\boldsymbol{\alpha} \cdot \boldsymbol{S_{id}} + \boldsymbol{\beta} \cdot \boldsymbol{S_{exp}},
\end{equation}
3DMM assumes that each face shares a similar structure that distributes around the average identity $\boldsymbol{\bar S_{id}}\in R^{3n}$. $ \boldsymbol{S_{id}} \in R^{3n\times 80}$, $\boldsymbol{S_{exp}} \in  R^{3n\times 29}$ are principal components representing the basis of identity and expression. $\boldsymbol{\alpha} \in R^{80}$ and $\boldsymbol{\beta} \in R^{29}$ are the use-specific coefficients estimated from the given image. In our implementation, the identity component comes from the Basel Face Model (BFW)~\cite{Blanz_1999}, whereas the expression comes from the FaceWarehouse database~\cite{Cao_2014}.

Synthesis is dependent on the 3DMM coefficients $\boldsymbol{\alpha}, \boldsymbol{\beta}$, the rigid translation $\boldsymbol{R}, \boldsymbol{t}$, an
d the camera projection matrix $\boldsymbol{\Pi}$. To reconstruct a 3D face model, we align corresponding 2D face landmarks with 3D landmarks on the bilinear face model using the pose normalization method~\cite{I.keme_2011}. All 3DMM parameters, defined as $\boldsymbol{\Theta}$, are then jointly estimated by using the following formula:
\begin{equation}
\begin{aligned}
\arg{\min_{\boldsymbol{\Theta}}}=\|\boldsymbol{\Pi}(\boldsymbol{RV}+\boldsymbol{t})-\boldsymbol{U}\|^2
+\rho_{1}\|\frac{\boldsymbol{\alpha}}{\boldsymbol{\xi_{id}}}\|^2+\rho_{2}\|\frac{\boldsymbol{\beta}}{\boldsymbol{\xi_{exp}}}\|^2,
\end{aligned}
\end{equation}
where $\boldsymbol{U}$ represents 2D face landmarks, and $\boldsymbol{V}$ represents corresponding vertices on the face model determined by 3DMM coefficients $\boldsymbol{\alpha}$, $\boldsymbol{\beta}$. $\rho_{1}$ and $\rho_{2}$ are positive weights of the regularization term to enforce the parameters to stay statistically close to the mean. $\boldsymbol{\xi_{id}}$ and $\boldsymbol{\xi_{exp}}$ are the standard deviations of shape and expression basis, respectively. We employ an occlusion-robust face alignment method~\cite{Bulat_2017_ICCV} to infer 68 face landmarks. All parameters are jointly solved via the Levenberg–Marquart algorithm~\cite{LM-optimization}.

\begin{figure}[t]
\begin{center}
    {\includegraphics[width=0.5\linewidth]{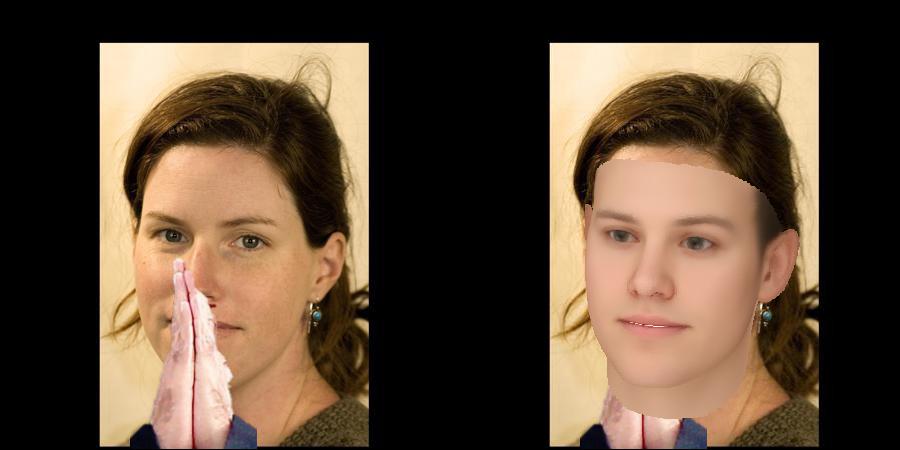}}~%
    {\includegraphics[width=0.5\linewidth]{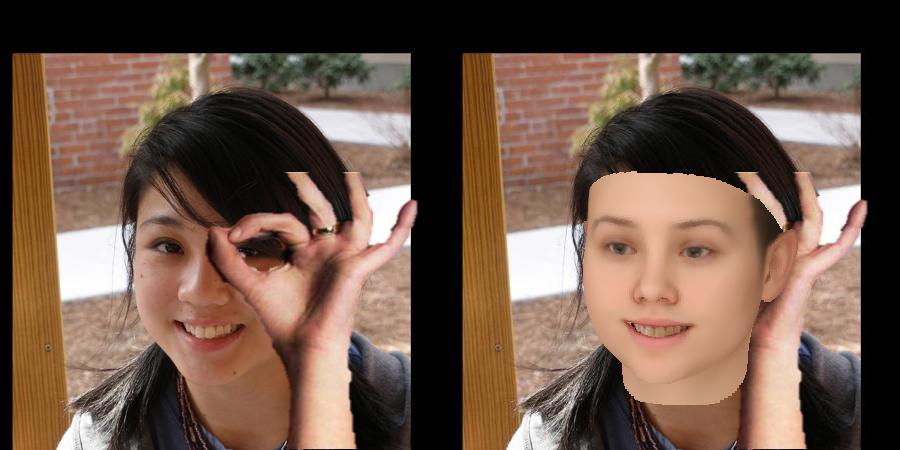}}~%
    \\ \vspace{0.5mm}
    {\includegraphics[width=0.5\linewidth]{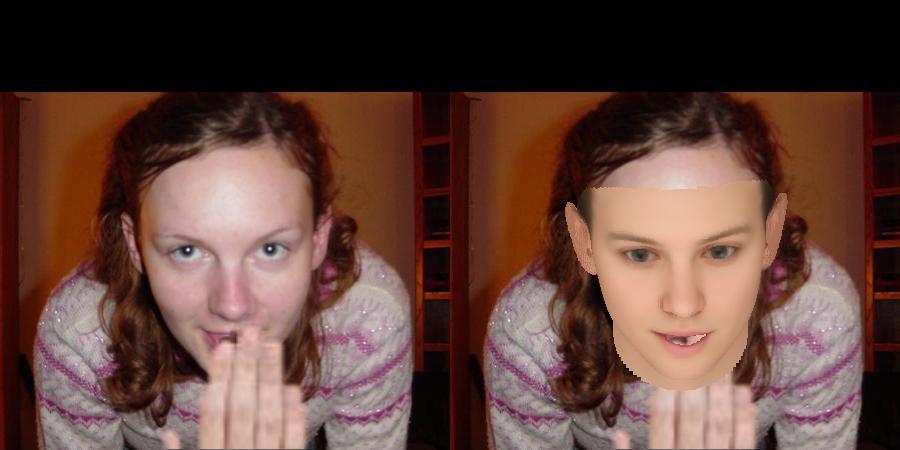}}~%
    {\includegraphics[width=0.5\linewidth]{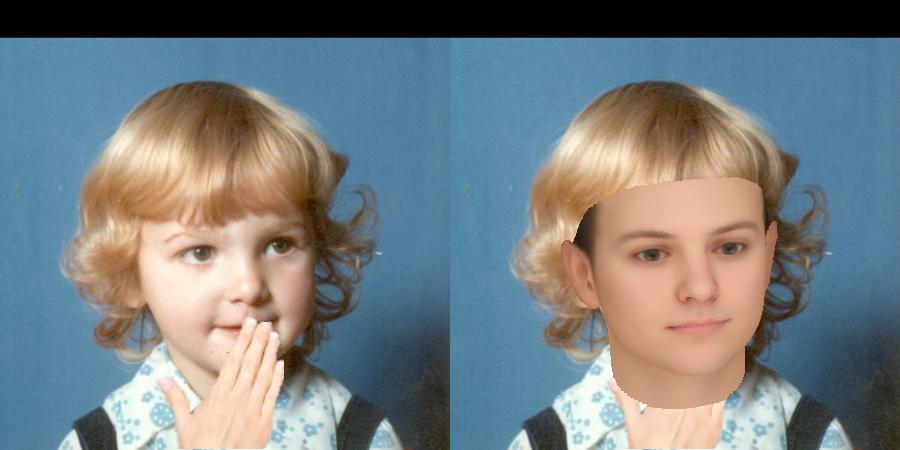}}~%
    \vspace{-3.6mm}
\end{center}
\caption{Occluded face image (left) and the 3DMM synthesis (right). }
    \label{fig:Result_3DMM}
\end{figure}

Based on the fitting result, the synthesis is generated as shown in Figure~\ref{fig:Result_3DMM}. The correspondence between pixels and triangles is computed by using Z buffering. Finally, the 3DMM is occlusion-free and can synthesize the appearance and pose of a face.

\section{Face De-occlusion using GAN}
The main framework of our model is a GAN that consists of a generator $G$, a global discriminator $D_g$, and a local discriminator $D_l$. The generator takes the occluded image and 3DMM synthesis as input to generate the occlusion-free image. Moreover, two discriminators $D_g$ and $D_l$ attempt to determine whether the generator output is a real face image or not. 3DMM not only serves as the prior but also provides a mask indicating the face region for the local discriminator. Additionally, a smoothness term is used to regularize $G$ to generate an image with fewer artifacts.

\subsection{Generator Module}
The generator G works as an auto encoder–decoder to remove face occlusion and construct the corrupted region. The occluded image $\boldsymbol{I}$, concatenated with the synthesis $\boldsymbol{I^s}$, is first mapped into the hidden feature through the encoder, which captures not only the variation of the known region but also the coarse geometric information of the occluded region. Then, the feature vector is fed into a decoder to generate an occlusion-free image.

Our encoder and decoder use modules of the form Convolution–BatchNorm –Relu and have the same architecture except for the input layer. We follow the encoder–decoder network designed in~\cite{isola2017image}, where a skip layer is used to preserve the low-level feature from the corresponding symmetrical layer. The skip collection allows combining the coarse geometry information from the downsampling path with the high-frequency features in the upsampling path to finally generate an occlusion-free image with good visual quality.

Even though 3DMM can synthesize the appearance and pose of a face, the generated image looks unrealistic and tends to lose all face details. To force the generator to output photo-realistic images, we adopt a pixel-wise $L_{1}$ reconstruction loss to penalize the output from the ground truth by using the following equation:
\begin{equation}
L_{gen}=|G(\boldsymbol{I},\boldsymbol{I^s})-\boldsymbol{I^g}|_1,
\end{equation}
where ~$\boldsymbol{I^g}$ is the ground truth, $\boldsymbol{I}$ is the occluded image, and $\boldsymbol{I^s}$ is the synthesis of 3DMM. As actual face images have various head poses, we avoid using symmetry loss as in the work of Yin~\cite{Yin_2017_ICCV}.

Despite the ability of the generator to reconstruct the occluded region with semantical contents, inconsistency occurs especially when the occlusion has a complex pattern. Thus, we use a total variation regularization to reduce the artifacts on the reconstructed region. We perform a $L_2$ minimization to the gradient of the generated image. The regularization is performed separately for each coordinate and then combined. The total variation regularization is commonly used in image noise removal using the following equation:
\begin{equation}
L_{tv}= |\nabla_{x} G(\boldsymbol{I},\boldsymbol{I^s})|_2 + |\nabla_{y} G(\boldsymbol{I},\boldsymbol{I^s})|_2
\end{equation}
However, this term tends to smoothen high-frequency details. Thus, we multiply a small weight to $L_{tv}$ to avoid oversmoothening.

Our generator can remove the occlusion and generate photo-realistic contents. However, recovering the expression in the occluded region is an ill-posed problem. Expression coefficients estimated from the occluded region can be arbitrary. Our generator relies on 3DMM for geometric information. Thus, we can edit face attributes by simply adjusting the 3DMM coefficients. Therefore, face de-occlusion and face attributes are integrated into one framework. The generator output of the occlusion-free image is consistent with the synthesis of 3DMM in geometry and demonstrates the solid effect of the 3DMM in regularizing the generation process.

\subsection{Discriminator Module}
Reconstruction loss tends to average all the details, thereby making the synthesized contents look blurry. Moreover, the generator only optimizes on the occluded region and cannot learn the relationship between pixels, which results in the generated contents being discontinuous with surroundings.

Recently, GAN consisting of generator and discriminator has been widely used for image synthesis. In this work, the generator synthesizes an occlusion-free face image, whereas the discriminator determines whether the generated face is real or not. The min-max optimization over generator and discriminator forces the model to synthesize images with better visual quality.

Our discriminator includes a local discriminator $D_l$ and a global discriminator $D_g$. The latter is used to determine the faithfulness of the entire image to enforce the generated region to become consistent with the surroundings. Considering our goal to reconstruct face geometry and texture synthesis on the image, we only rely on the face region. Thus, we enforce the optimization of the local discriminator in the face region. The mask $\mathcal{M}$ used for the local discriminator is a projected silhouette from the 3DMM indicating the face region. Compared with the global discriminator, the local module enhances details in the face region with well-defined boundaries and less noise.

By combining the local module with the global module, we not only guarantee the statistical consistency of the generated face region with its surroundings but also encourage the recovered face region to become highly informative. To train these two discriminators, the following objectives are minimized:
\begin{equation}
\begin{aligned}
L_{D_g}=&-\mathbb{E}_{\boldsymbol{I^g} \in R}\log D_g(\boldsymbol{I^g)}\\
&-\mathbb{E}_{\boldsymbol{I^s} \in K}\log(1- D_g(G(\boldsymbol{I},\boldsymbol{I^s}))),
\end{aligned}
\end{equation}
\begin{equation}
\begin{aligned}
L_{D_l}=&-\mathbb{E}_{\boldsymbol{I^g} \in R}\log D_l(\mathcal{M}\odot(\boldsymbol{I^g}))\\
        &-\mathbb{E}_{\boldsymbol{I^s} \in K} \log(1- D_l(\mathcal{M}\odot G(\boldsymbol{I},\boldsymbol{I^s}))),
\end{aligned}
\end{equation}
where the $\odot$ denotes the element-wise multiplication and $R$ and $K$ are real and generated image sets, respectively. Our two discriminators have similar network structures that consist of seven convolution layers. After the last layer, a convolution is mapped to one-dimensional output, followed by a sigmoid function. The outputs of the discriminators determine whether the probability of the input is real or generated.

In addition, $G$ attempts to fool the two discriminators in identifying the generated image as real by minimizing the following loss:
\begin{equation}
\begin{aligned}
L_{adv_l}=&-\mathbb{E}_{\boldsymbol{I^s} \in K} \log(D_l(\mathcal{M}\odot G(\boldsymbol{I},\boldsymbol{I^s})))\\
L_{adv_g}=&-\mathbb{E}_{\boldsymbol{I^s} \in K} \log(D_g(G(\boldsymbol{I},\boldsymbol{I^s})))
\end{aligned}
\end{equation}

\subsection{Objective Function}
To summarize, the final loss for our proposed 3DMM-conditioned GAN is represented as a weighted sum of the aforementioned losses:
\begin{equation}
\begin{aligned}
L=&\lambda_{1} L_{gen}+\lambda_{2} L_{tv}+\lambda_{3} L_{adv_g}+\\
  &\lambda_{4} L_{adv_l}+\lambda_{5} L_{D_l}+\lambda_{6} L_{D_g}
\end{aligned}
\end{equation}
Weights $\lambda_1$, $\lambda_2$, $\lambda_3$, $\lambda_4$, $\lambda_5$, and $\lambda_6$ are used to balance different terms.

\section{Experimental Result}

\subsection{Dataset}
Occluded images, 3DMM synthesis, and corresponding occlusion-free images are needed to train our face de-occlusion model. Owing to the difficulty in collecting sufficient occluded face images with their corresponding occlusion-free images, we train our model on our synthesized dataset. The datasets used for training and testing are introduced in the following.

\vspace{-2mm}
\paragraph{300W-3D}
This dataset consists of 7,700 300-W~\cite{Sagonas_2013_CVPR} samples with the fitted 3DMM parameters and 68 face landmarks for each sample. All images in 300-W are real photos and cover variations in pose, illumination, background, and image quality.

\vspace{-2mm}
\paragraph{AFLW2000-3D}
This dataset consists of the first 2,000 images in AFLW ~\cite{AFLW}. Similar to 300W-3D, 3DMM parameters and 68 landmarks are provided for each image.

\vspace{-2mm}
\paragraph{CelebA~\cite{Liu_2015_ICCV}}
This dataset consists of 202,599 celebrity images with each image cropped and roughly aligned by the positions of the two eyes. We select occluded faces in this dataset for testing.

\begin{figure}[t]
\begin{center}
    {\includegraphics[width=0.2\linewidth]{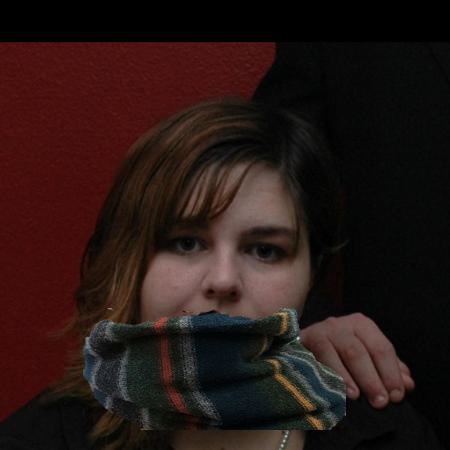}}~%
    {\includegraphics[width=0.2\linewidth]{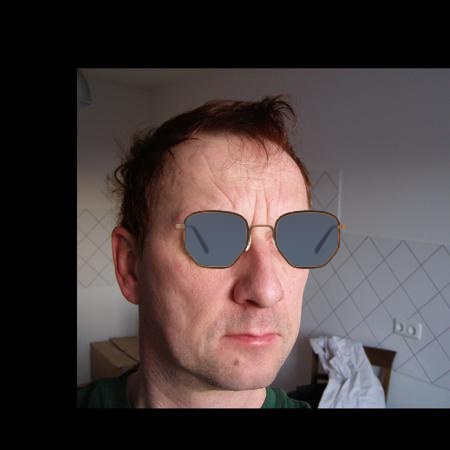}}~%
    {\includegraphics[width=0.2\linewidth]{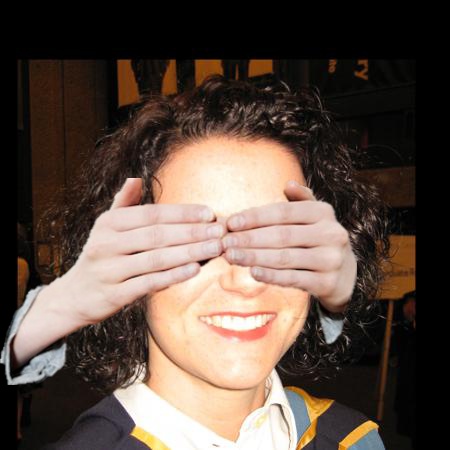}}~%
    {\includegraphics[width=0.2\linewidth]{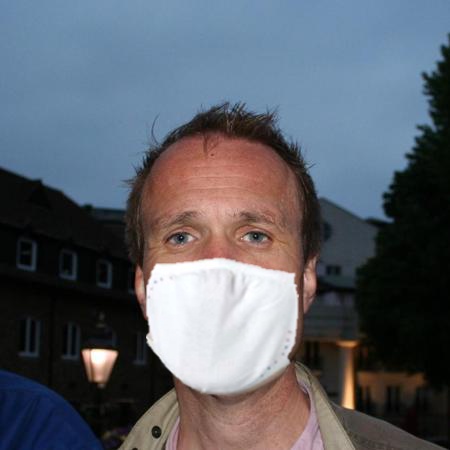}}~%
    {\includegraphics[width=0.2\linewidth]{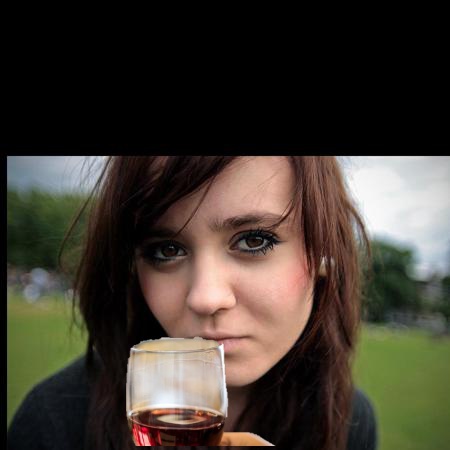}}~%

    \vspace{0.5mm}
    {\includegraphics[width=0.2\linewidth]{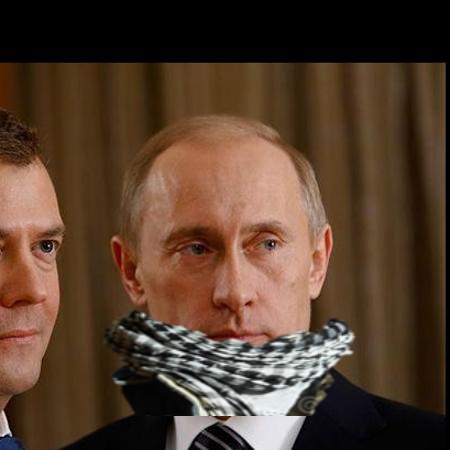}}~%
    {\includegraphics[width=0.2\linewidth]{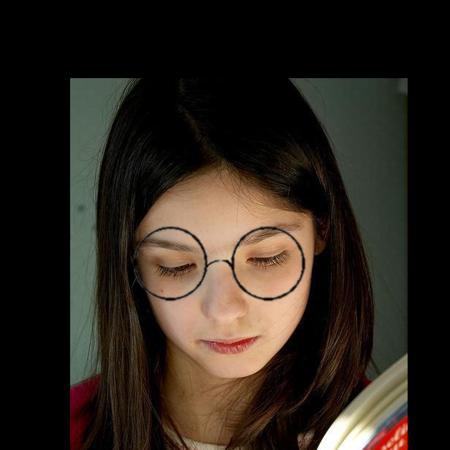}}~%
    {\includegraphics[width=0.2\linewidth]{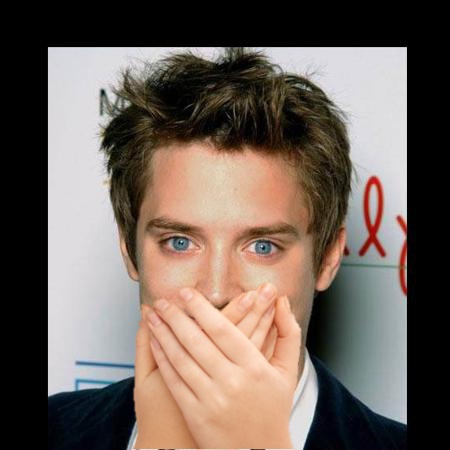}}~%
    {\includegraphics[width=0.2\linewidth]{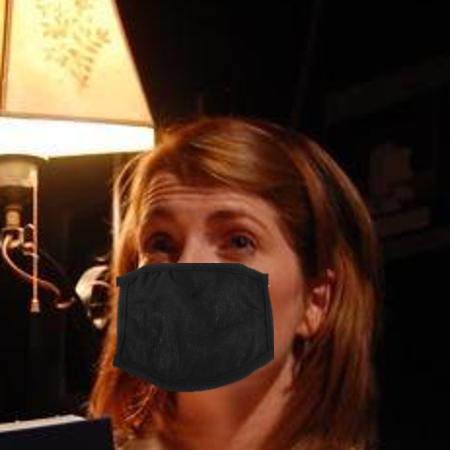}}~%
    {\includegraphics[width=0.2\linewidth]{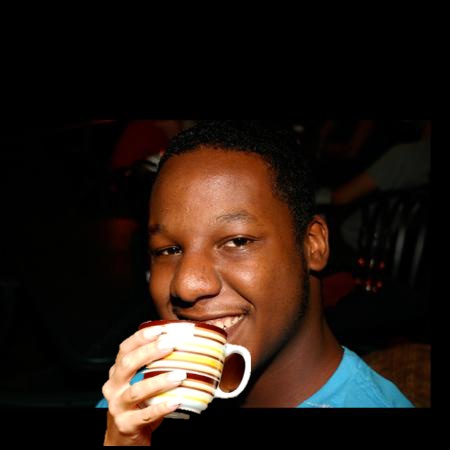}}~%
    \vspace{-3.6mm}
\end{center}
\caption{Samples of our synthesized dataset for training. Occlusions are located semantically based on the face landmarks.}
    \label{fig:Result_dataset}
\end{figure}

\vspace{-2mm}
\paragraph{} We synthesize occlusions caused by six common objects on occlusion-free faces in 300W-3D and AFLW2000-3D. These objects include masks, eyeglasses, sunglasses, cups, scarves, and hands. We layer these occlusions on the specific location of the face with reference to the face landmarks. Figure~\ref{fig:Result_dataset} shows examples of the occluded faces generated using this approach. All occlusions are semantically located on the face to augment the reality of our dataset. Then, we generate the synthesized image of 3DMM for every training sample by using 3DMM coefficients and camera pose provided by 300W-3D and AFLW2000-3D. We synthesize a dataset with a total of 134,233 occluded images. All faces in the dataset are resized $256\times256$ and with head poses varying from $60^\circ$ to $60^\circ $. We select 132,233 images for training and 2,000 images for testing. Random cropping and horizontal flipping are used in data augmentation to avoid overfitting. Besides using the synthesized images, we test our model on real images, which consist of the occluded images from the aforementioned three datasets.

\begin{figure*}[t]
\floatsetup[figure]{style=plain,subcapbesideposition=top}
\begin{center}
    \hfill
    \sidesubfloat[]
    {\includegraphics[width=0.12\linewidth]{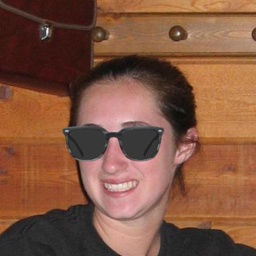}}~%
    {\includegraphics[width=0.12\linewidth]{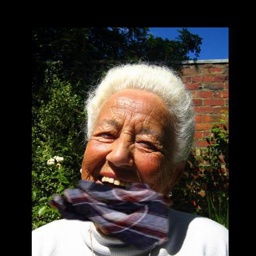}}~%
    {\includegraphics[width=0.12\linewidth]{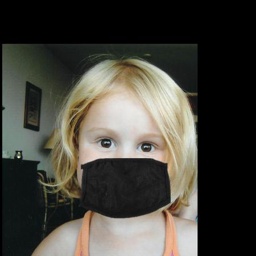}}~%
    {\includegraphics[width=0.12\linewidth]{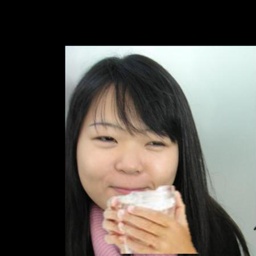}}~%
    {\includegraphics[width=0.12\linewidth]{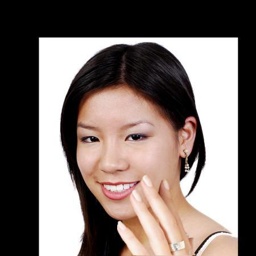}}~%
    {\includegraphics[width=0.12\linewidth]{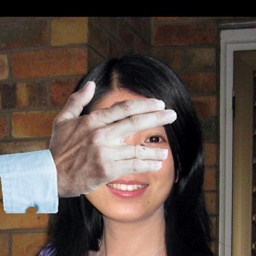}}~%
    {\includegraphics[width=0.12\linewidth]{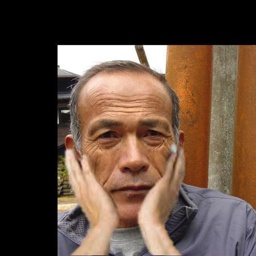}}~%
    {\includegraphics[width=0.12\linewidth]{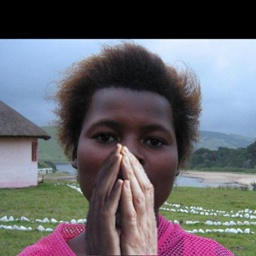}} \\~%
    \vspace{-3.5mm}
    \hfill
    \sidesubfloat[]
    {\includegraphics[width=0.12\linewidth]{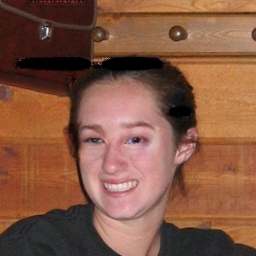}}~%
    {\includegraphics[width=0.12\linewidth]{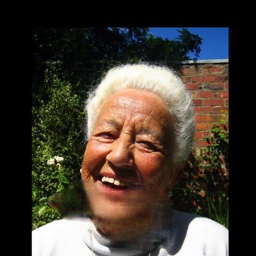}}~%
    {\includegraphics[width=0.12\linewidth]{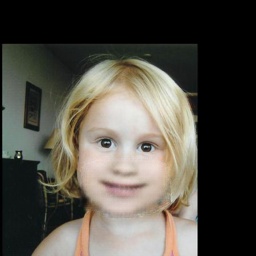}}~%
    {\includegraphics[width=0.12\linewidth]{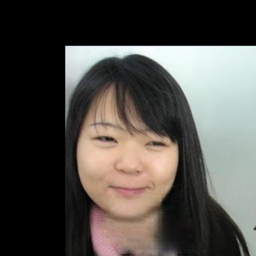}}~%
    {\includegraphics[width=0.12\linewidth]{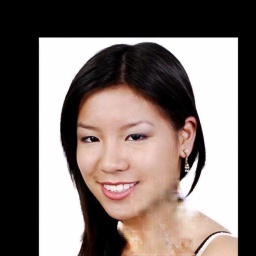}}~%
    {\includegraphics[width=0.12\linewidth]{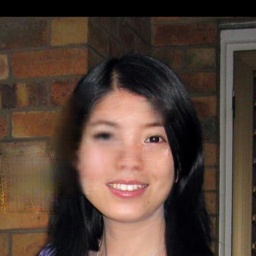}}~%
    {\includegraphics[width=0.12\linewidth]{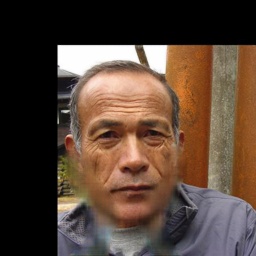}}~%
    {\includegraphics[width=0.12\linewidth]{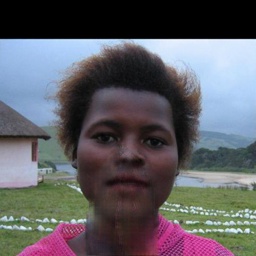}} \\~%
    \hfill
    \vspace{0.6mm}
    \sidesubfloat[]
    {\includegraphics[width=0.12\linewidth]{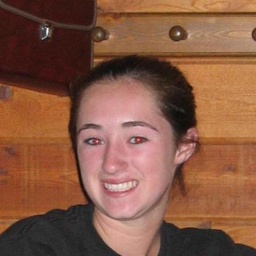}}~%
    {\includegraphics[width=0.12\linewidth]{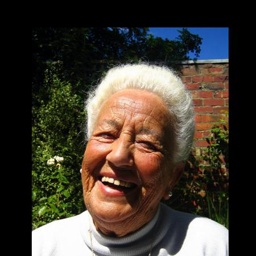}}~%
    {\includegraphics[width=0.12\linewidth]{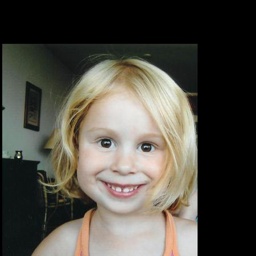}}~%
    {\includegraphics[width=0.12\linewidth]{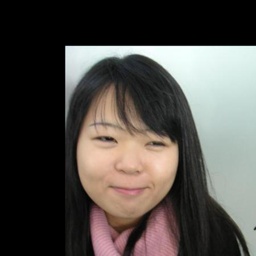}}~%
    {\includegraphics[width=0.12\linewidth]{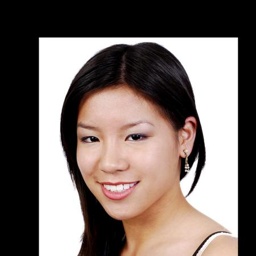}}~%
    {\includegraphics[width=0.12\linewidth]{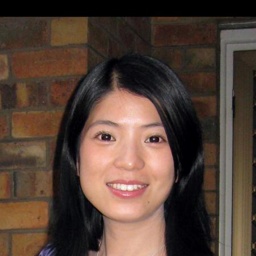}}~%
    {\includegraphics[width=0.12\linewidth]{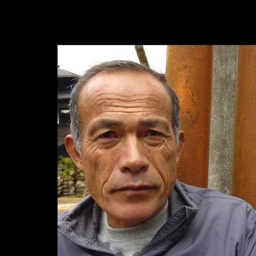}}~%
    {\includegraphics[width=0.12\linewidth]{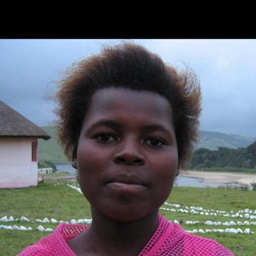}}~%
    \vspace{-3.6mm}
\end{center}
\caption{Face de-occlusion results on the synthetic dataset. (a) Image with occlusion. (b) De-occlusion result. (c) Original image (ground truth).}
    \label{fig:Result_synthetic}
\end{figure*}
\begin{figure*}[t]
\floatsetup[figure]{style=plain,subcapbesideposition=top}
\begin{center}
    \hfill
    \sidesubfloat[]{\includegraphics[width=0.12\linewidth]{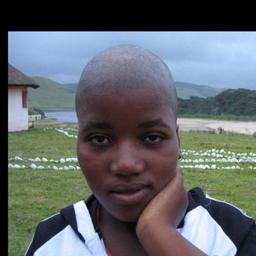}}~%
    {\includegraphics[width=0.12\linewidth]{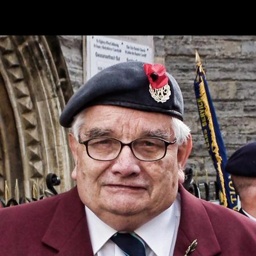}}~%
    {\includegraphics[width=0.12\linewidth]{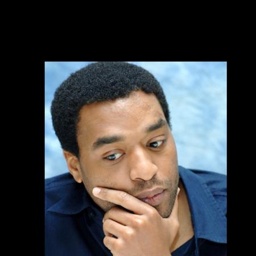}}~%
    {\includegraphics[width=0.12\linewidth]{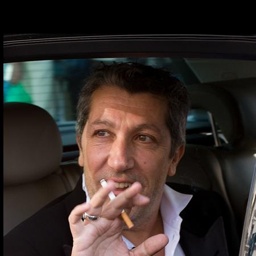}}~%
    {\includegraphics[width=0.12\linewidth]{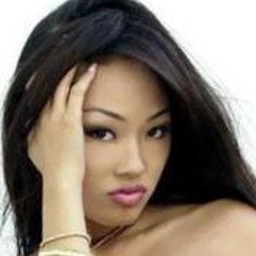}}~%
    {\includegraphics[width=0.12\linewidth]{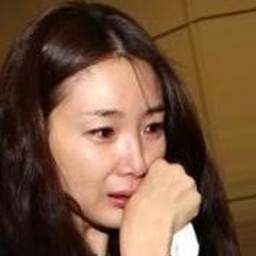}}~%
    {\includegraphics[width=0.12\linewidth]{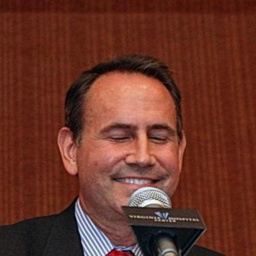}}~%
    {\includegraphics[width=0.12\linewidth]{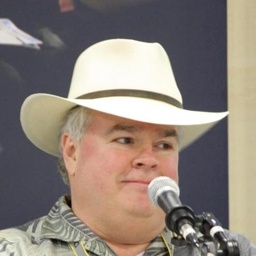}}~%
    \hfill
    \vspace{0.6mm}
    \sidesubfloat[]{\includegraphics[width=0.12\linewidth]{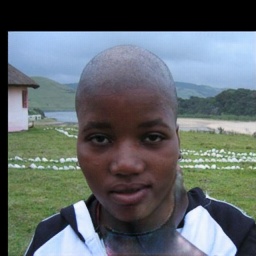}}~%
    {\includegraphics[width=0.12\linewidth]{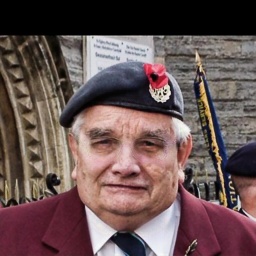}}~%
    {\includegraphics[width=0.12\linewidth]{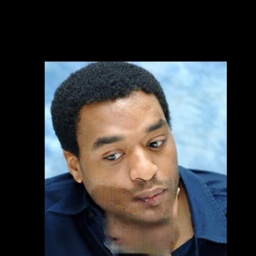}}~%
    {\includegraphics[width=0.12\linewidth]{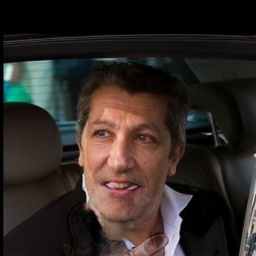}}~%
    {\includegraphics[width=0.12\linewidth]{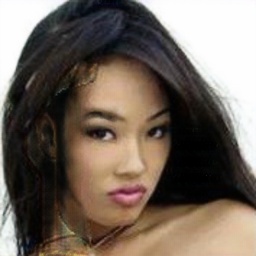}}~%
    {\includegraphics[width=0.12\linewidth]{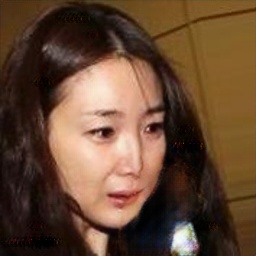}}~%
    {\includegraphics[width=0.12\linewidth]{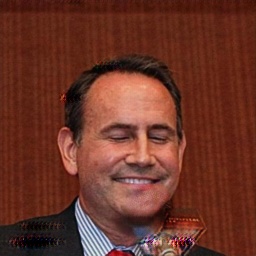}}~%
    {\includegraphics[width=0.12\linewidth]{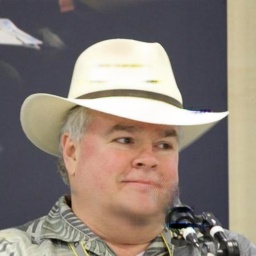}}~%
    \vspace{-3.6mm}
\end{center}
\caption{Face de-occlusion results on the real dataset. (a) Image with occlusion. (b) De-occlusion result.}
    \label{fig:Result_realData}
\end{figure*}
\subsection{Implementation Details}
We train our network with batch size 5 and utilize Adam optimizer. Instead of jointly training all modules, we gradually add them. In the first stage, we train the generator and global discriminator with a learning rate of $0.0002$ for 100 epochs. In the second stage, we add the local discriminator and remove the total variation regularization to finetune the network with a learning rate of $0.00005$, and train another 10 epochs. During training, we set the value of $\lambda_1$, $\lambda_2$, $\lambda_3$, $\lambda_4$, $\lambda_5$, $\lambda_6$ as 10, $10^{-5}$, 1, 1, 1, and 1, respectively. In the testing stage, only the generation module is required. The entire training procedure takes approximately 4d on a single GeForce GTX 1080Ti GPU. In the testing stage, a $256\times256$ color image can be processed in under a second.

\subsection{Face De-occlusion}
\paragraph{Qualitative Result}
Figure~\ref{fig:Result_synthetic} and Figure~\ref{fig:Result_realData} show the face de-occlusion results on the synthetic and real images, respectively. Note that the identities in the test dataset are separated from the training dataset. As shown in Figure~\ref{fig:Result_synthetic}(b) and Figure~\ref{fig:Result_realData}(a), test images have various types of occlusion at arbitrary locations. The results show that the proposed method successfully removes the occlusion and generates a photorealistic de-occluded image for both synthetic and real data even when a significant portion of the face region is occluded. As shown in the last two examples in Figure~\ref{fig:Result_realData}, the proposed method model removes not only occlusions similar to that in our training dataset but also those that do not exist in our dataset (no occlusion with microphones in the training dataset). The result confirms that the proposed 3DMM-conditioned face de-occlusion model can remove different types of occlusions with challenging conditions including various head poses and illumination.

Failure occurs when more than one type of occlusion exists on the face and when the occlusion is located out of the synthesis range of the 3DMM parametric space, {\em e.g.}, hands above the forehead.

\begin{figure}[t]
\begin{center}
    \hfill
    {\includegraphics[width=0.25\linewidth]{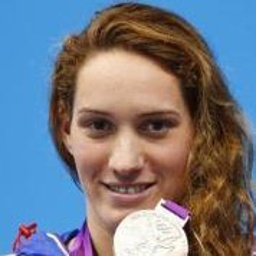}}~%
    {\includegraphics[width=0.25\linewidth]{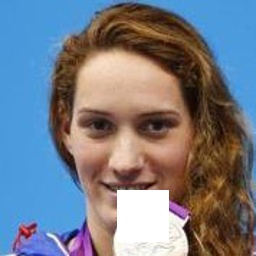}}~%
    {\includegraphics[width=0.25\linewidth]{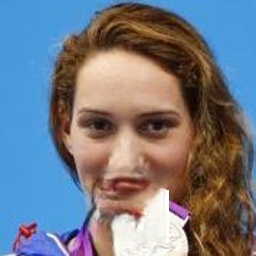}}~%
    {\includegraphics[width=0.25\linewidth]{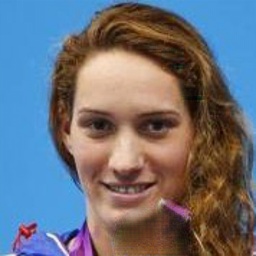}}~\\%
\vspace{0.5mm}
    {\includegraphics[width=0.25\linewidth]{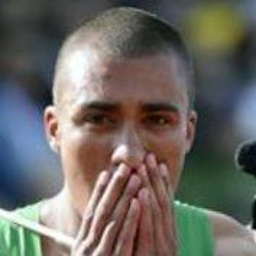}}~%
    {\includegraphics[width=0.25\linewidth]{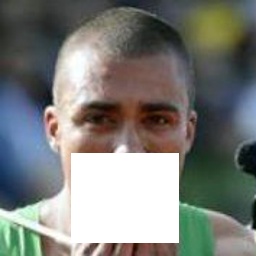}}~%
    {\includegraphics[width=0.25\linewidth]{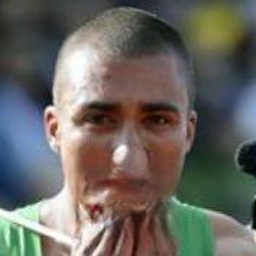}}~%
    {\includegraphics[width=0.25\linewidth]{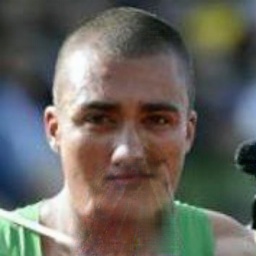}}~\\%
\vspace{-3.0mm}
    \subfloat[]{\includegraphics[width=0.25\linewidth]{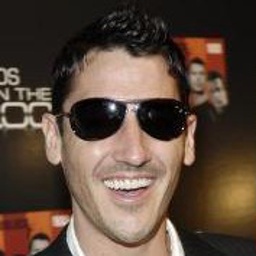}}~%
    \subfloat[]{\includegraphics[width=0.25\linewidth]{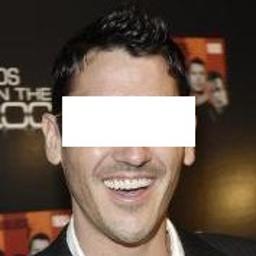}}~%
    \subfloat[]{\includegraphics[width=0.25\linewidth]{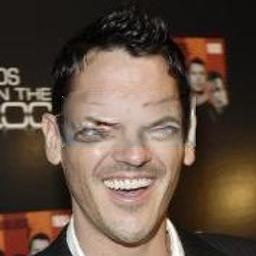}}~%
    \subfloat[]{\includegraphics[width=0.25\linewidth]{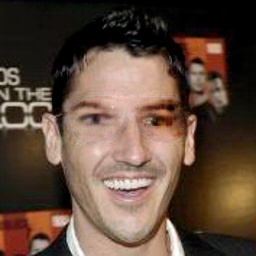}}~%

\end{center}
\vspace{-3.6mm}
\caption{Comparison result with the state-of-the-art face inpainting algorithm~\cite{Yu_2018_CVPR}. (a) Input image with occlusion. (b) Masking the occluded region. (c) De-occlusion result with the inpainting algorithm. (d) De-occlusion result with the proposed algorithm. }
    \label{fig:Result_comparison}
\end{figure}
\begin{table}
\caption{Quantitative evaluation for different types of occlusion.}
\vspace{2mm}
\centering
\begin{tabular}{c|c|c}
\toprule
Type of occlusion               & PSNR     & SSIM  \\
\hline
Lower face                      & 27.3228  & 0.9615  \\
Upper face                      & 34.0024  & 0.9860  \\
Left/right half of face         & 28.7785  & 0.9659 \\
Three quarters of face          & 22.1680  & 0.8967 \\
\hline
\end{tabular}
  \label{table:Quantitative evaluation}
\end{table}
\paragraph{Quantitative Result}
To quantitatively measure the de-occlusion performance, two popular metrics, {\em i.e.}, PSNR and SSIM, are evaluated on the de-occlusion result of the synthetic dataset and listed in Table~\ref{table:Quantitative evaluation}. The performance of our model slightly drops when more than half of the face is occluded, which is expected as a large occlusion size indicates uncertainty in pixel values. The model also shows better de-occlusion performance on the upper face than the lower face because occlusions on the lower face region have complex patterns, such as different scarves and cups.

\paragraph{Comparison}
The goal of the proposed face de-occlusion model is to remove occlusions on face images and recover the missing region. As the goal of the previous methods~\cite{Wring_2009, Cheng_2015,Zhao_2018} is completely different (face recognition with a low-resolution grayscale image), we do not compare our results with theirs. Instead, we compare our approach with the recent state-of-art face inpainting method~\cite{Yu_2018_CVPR} because the recent face inpainting method shows potential application in removing occlusions and reconstructing de-occluded face regions. As the method proposed by Yu~\cite{Yu_2018_CVPR} is only trained on the {\it CelebA} dataset, we conduct the experiment on occluded images from that dataset to be fair. First, the occluded region is masked with the provided pattern and the inpainting algorithm is applied to reconstruct the masked region. As shown in Figure~\ref{fig:Result_comparison}, the inpainting algorithm does not work effectively on face images. Face inpainting is usually utilized on a well-aligned dataset, thereby failing to generate the semantic contents on difficult cases, such as posed face and complex backgrounds. On the contrary, the proposed method can automatically remove occlusions without any preprocessing on the occluded region while showing significantly better results.


\begin{figure}[t]
\begin{center}
    {\includegraphics[width=0.16\linewidth]{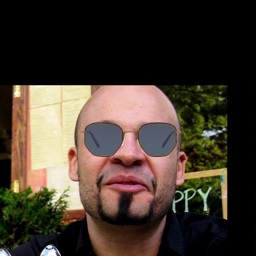}}~%
    {\includegraphics[width=0.16\linewidth]{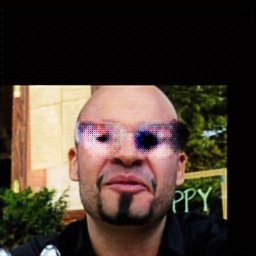}}~%
    {\includegraphics[width=0.16\linewidth]{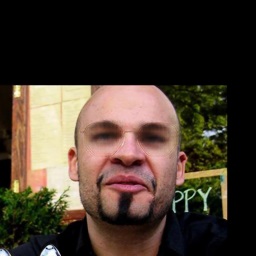}}~%
    {\includegraphics[width=0.16\linewidth]{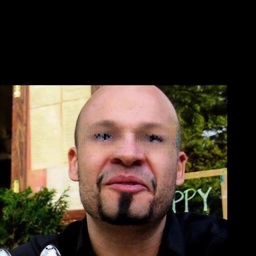}}~%
    {\includegraphics[width=0.16\linewidth]{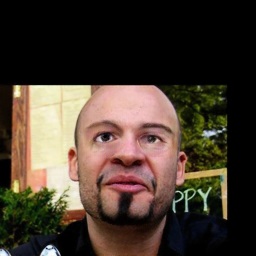}}~%
    {\includegraphics[width=0.16\linewidth]{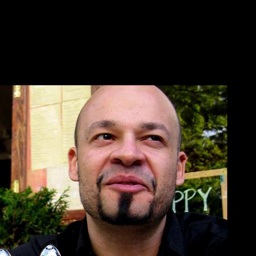}}~%

    {\includegraphics[width=0.16\linewidth]{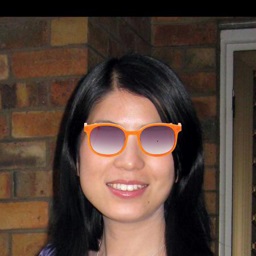}}~%
    {\includegraphics[width=0.16\linewidth]{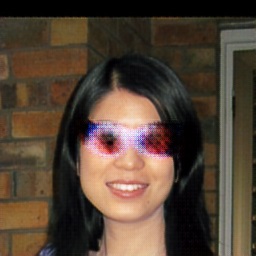}}~%
    {\includegraphics[width=0.16\linewidth]{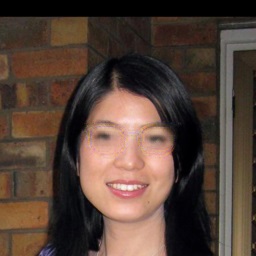}}~%
    {\includegraphics[width=0.16\linewidth]{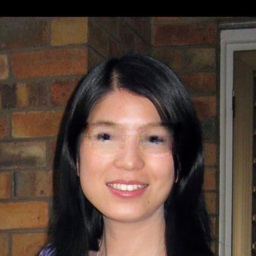}}~%
    {\includegraphics[width=0.16\linewidth]{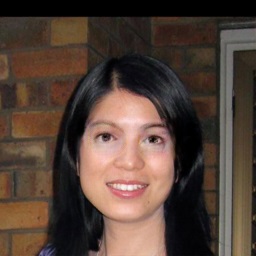}}~%
    {\includegraphics[width=0.16\linewidth]{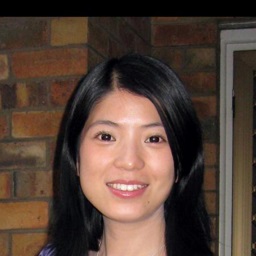}}~%

\vspace{-3.2mm}
    \subfloat[]{\includegraphics[width=0.16\linewidth]{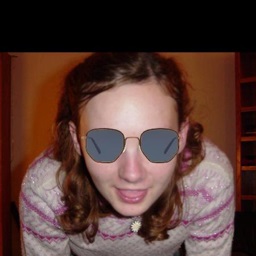}}~%
    \subfloat[]{\includegraphics[width=0.16\linewidth]{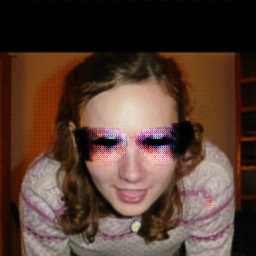}}~%
    \subfloat[]{\includegraphics[width=0.16\linewidth]{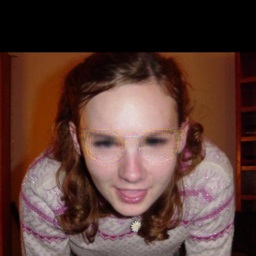}}~%
    \subfloat[]{\includegraphics[width=0.16\linewidth]{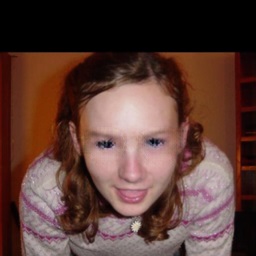}}~%
    \subfloat[]{\includegraphics[width=0.16\linewidth]{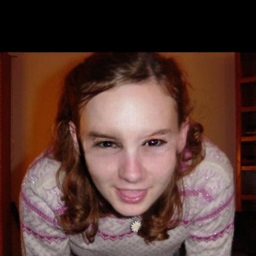}}~%
    \subfloat[]{\includegraphics[width=0.16\linewidth]{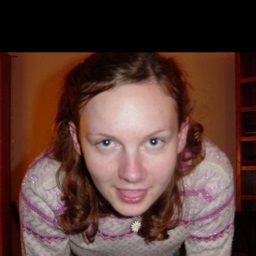}}~%

\end{center}
\vspace{-3.6mm}
\caption{Comparison result under different setting. (a) Occluded face image. (b) Without 3DMM synthesis. (c) With generator only. (d) With global discriminator. (e) With global and local discriminators. (f) Ground truth.}
    \label{fig:Result_ablate}
\end{figure}
\paragraph{Ablation Study}
To validate the effects of the 3DMM synthesis, we train the other variants with similar hyperparameters but different settings and compare the performance. We remove 3DMM, global discriminator, and local discriminator in turn.
Without 3DMM, the network generates noisy outputs or fails to generate informative results. The result is sensible because generator with only pixel-wise reconstruction loss is too weak to learn the representation of the face geometry from a challenging face dataset. Note that, in face de-occlusion, we have to find and restore the occluded region while handling the pose variation simultaneously which is a serious ill-posed problem. By using 3DMM as a prior, the ill-posedness can be alleviated and de-occlusion on images with various head poses can be performed properly.
Without discriminators, the model can generate images with semantical contents but artifacts remain on the recovered region. With only the global discriminator, the result looks sharp and coherent, but lacks details in the eyes. With the combined global and local discriminators, the face de-occlusion results look visually realistic. The visual comparison results are summarized in Figure~\ref{fig:Result_ablate}.

\begin{figure}[t]
\begin{center}
    {\includegraphics[width=0.2\linewidth]{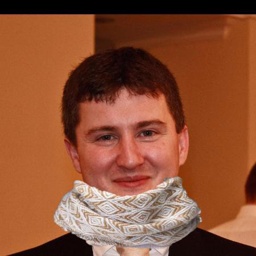}}~%
    {\includegraphics[width=0.2\linewidth]{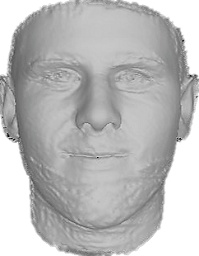}}~%
    {\includegraphics[width=0.2\linewidth]{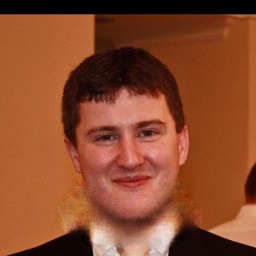}}~%
    {\includegraphics[width=0.2\linewidth]{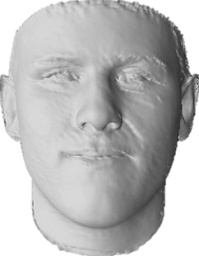}}~%
    {\includegraphics[width=0.2\linewidth]{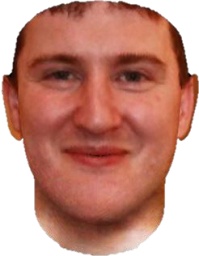}}~%

    {\includegraphics[width=0.2\linewidth]{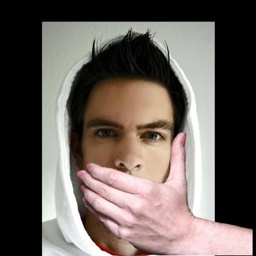}}~%
    {\includegraphics[width=0.2\linewidth]{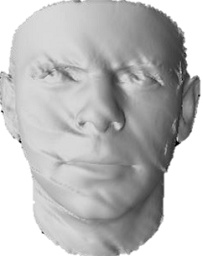}}~%
    {\includegraphics[width=0.2\linewidth]{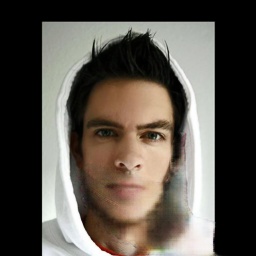}}~%
    {\includegraphics[width=0.2\linewidth]{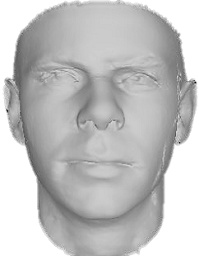}}~%
    {\includegraphics[width=0.2\linewidth]{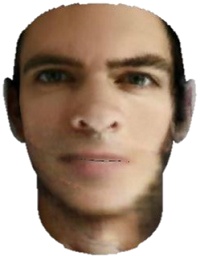}}~%

    {\includegraphics[width=0.2\linewidth]{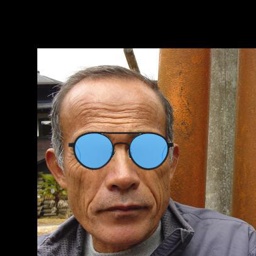}}~%
    {\includegraphics[width=0.2\linewidth]{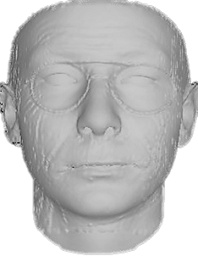}}~%
    {\includegraphics[width=0.2\linewidth]{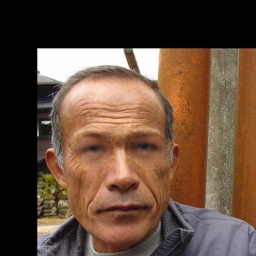}}~%
    {\includegraphics[width=0.2\linewidth]{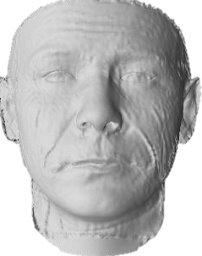}}~%
    {\includegraphics[width=0.2\linewidth]{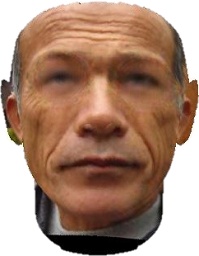}}~%

    \subfloat[]{\includegraphics[width=0.2\linewidth]{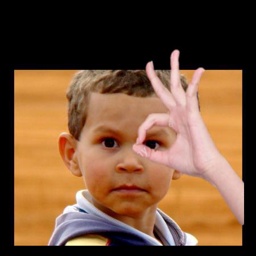}}~%
    \subfloat[]{\includegraphics[width=0.2\linewidth]{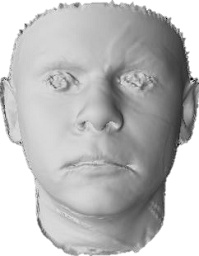}}~%
    \subfloat[]{\includegraphics[width=0.2\linewidth]{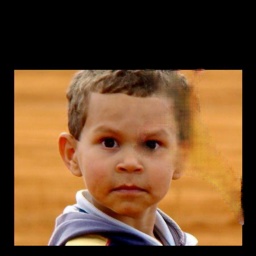}}~%
    \subfloat[]{\includegraphics[width=0.2\linewidth]{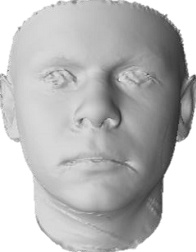}}~%
    \subfloat[]{\includegraphics[width=0.2\linewidth]{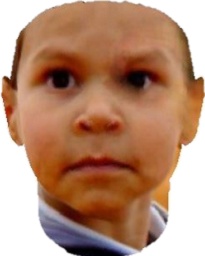}}~%
    \vspace{-3.6mm}
\end{center}
\caption{Comparison results of detailed face reconstruction from occluded image and de-occluded image. (a) Occluded face image. (b) Face reconstruction result with (a). (c) De-occluded image. (d) Face reconstruction result with (c). (e) 3D face model with de-occluded texture mapping.}
\label{fig:Result_sfs}
\end{figure}
\subsection{3D Face Reconstruction}
As our motivation is face de-occlusion for 3D face reconstruction, we conduct the experiments to investigate the effect of our face de-occlusion model on 3D face reconstruction. In our experiment, coarse 3D face model and detailed face geometry are reconstructed with the de-occluded face image. For this purpose, conventional landmark-based 3DMM fitting is conducted first and the shape-from-shading~(SfS) method is employed to enhance details on the coarse face model~\cite{Kemelmacher_2011, Richardson_2017_CVPR, Jiang_2018}. Based on the assumption that Lambertian reflection on the face exists, the intensity formation of the face image can be represented as follows:
\begin{equation}
    \boldsymbol{I}(x,y)=\boldsymbol{\rho}(x,y)\boldsymbol{\vec{l}}~\boldsymbol{Y}(\boldsymbol{\vec{n}}(x,y)),
\end{equation}
where $\boldsymbol{Y}(\boldsymbol{\vec{n}}(x,y))$ is the second-order spherical harmonics~\cite{spherical}, $\boldsymbol{\vec{l}}$ represents lighting coefficients, $\boldsymbol{\rho}(x,y)$, and $\boldsymbol{\vec{n}}(x,y)$ are the albedo and normal vector at pixel$(x, y)$, respectively. Following the work of Kemelmacher~\cite{Kemelmacher_2011}, we estimate lighting $\boldsymbol{\vec{l}}$, albedo $\boldsymbol{\rho}(x,y)$, and normal vector $\boldsymbol{\vec{n}}(x,y)$ in turn. Then, the estimated normal vector $\boldsymbol{\vec{n}}(x,y)$ is integrated to recover the detailed face geometry.

Figure~\ref{fig:Result_sfs} shows the comparison results of detailed 3D face reconstruction using face images with occlusion and de-occlusion. The results demonstrate that occlusion causes significant noise on the face geometry. Note that the geometry is not completely corrupted because the shape is still controlled by 3DMM. On the contrary, by removing occlusions using the proposed method, both face geometry and textured 3D face model are reconstructed correctly.

\begin{figure}[t]
\begin{center}
    {\includegraphics[width=0.2\linewidth]{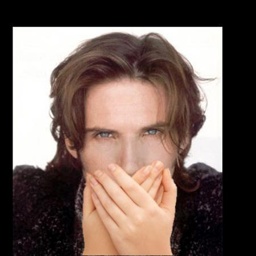}}~%
    {\includegraphics[width=0.2\linewidth]{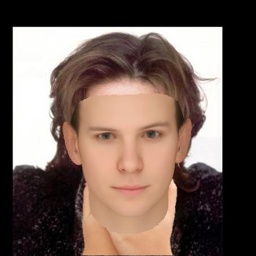}}~%
    {\includegraphics[width=0.2\linewidth]{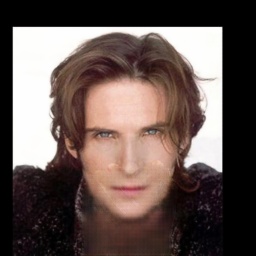}}~%
    {\includegraphics[width=0.2\linewidth]{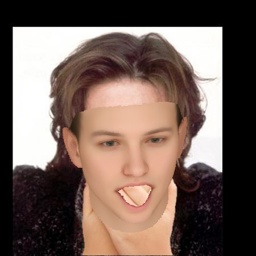}}~%
    {\includegraphics[width=0.2\linewidth]{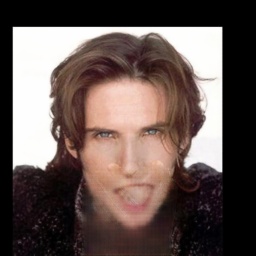}}~%

    \vspace{0.5mm}
    {\includegraphics[width=0.2\linewidth]{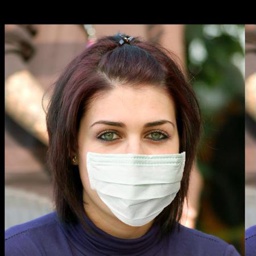}}~%
    {\includegraphics[width=0.2\linewidth]{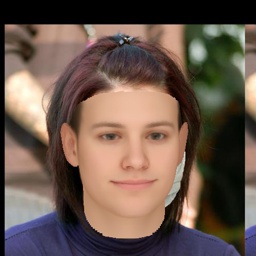}}~%
    {\includegraphics[width=0.2\linewidth]{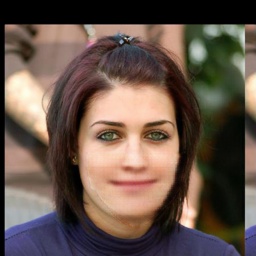}}~%
    {\includegraphics[width=0.2\linewidth]{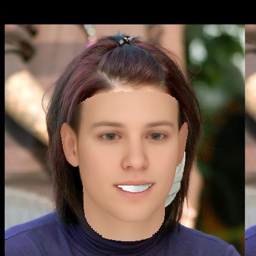}}~%
    {\includegraphics[width=0.2\linewidth]{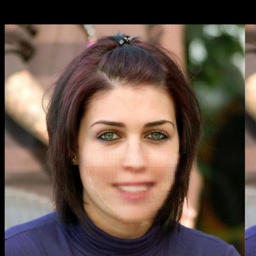}}~%


	\vspace{-3mm}
    \subfloat[]{\includegraphics[width=0.2\linewidth]{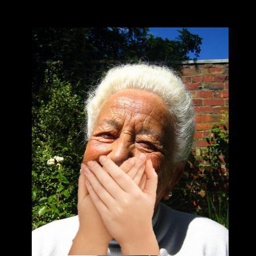}}~%
    \subfloat[]{\includegraphics[width=0.2\linewidth]{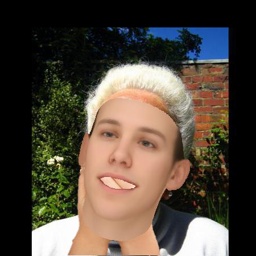}}~%
    \subfloat[]{\includegraphics[width=0.2\linewidth]{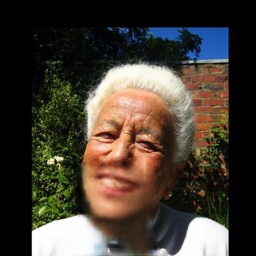}}~%
    \subfloat[]{\includegraphics[width=0.2\linewidth]{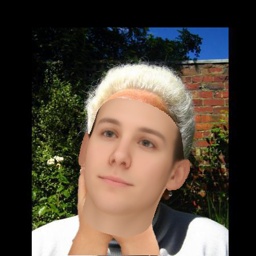}}~%
    \subfloat[]{\includegraphics[width=0.2\linewidth]{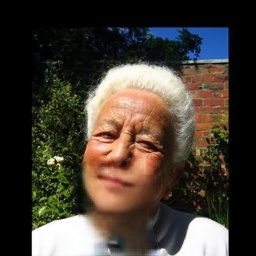}}~%
    \vspace{-3.6mm}
\end{center}
\caption{Results of face attribute editing. (a) Occluded face image. (b)(d) 3DMM synthesis with different expression coefficients. (c) Generated face guided by (b). (e) Generated face guided by (d).}
    \label{fig:Result_attribute}
\end{figure}
\subsection{Face Attributes Manipulation}
The proposed face de-occlusion can be applied to future studies, such as face editing and face recognition, to improve performance. In this further experiment, we show the application of the proposed model in face attribute editing.

As the proposed generative model recovers the occluded face guided by the 3DMM synthesis, it allows face attribute editing by simply adjusting the 3DMM coefficients to any desired one. Therefore, the proposed model holds potential application in face editing to generate a novel portrait, as shown in Figure~\ref{fig:Result_attribute}. Given the same occluded image, we can modify the attribute of the generated face by changing 3DMM expression coefficients as shown in Figure~\ref{fig:Result_attribute}(b) and (d).

\section{Conclusion}

In this paper, we proposed a 3DMM-conditioned GAN framework to remove face occlusion and restore the occluded region. To the best of our knowledge, this study is the first to explore the use of 3DMM in face de-occlusion on challenging dataset. Experimental results show that our face de-occlusion model can remove face occlusion on synthetic and real images. The proposed method not only removes the occlusion but also reconstructs the correct 3D face model without occluded texture. Furthermore, our method allows face attribute editing by simply modifying the 3DMM coefficients.

\section*{Acknowledgement}

This work was supported by the National Research Foundation of Korea~(NRF)~grant funded by the Korea government~(MSIT)~(No. NRF-2019R1A2C1006706).

{\small
\bibliographystyle{ieee_fullname}
\bibliography{egbib}
}
\end{document}